%% file: main.tex
\documentclass[runningheads]{llncs}
\usepackage{graphicx}
\usepackage{makecell}
\usepackage{amsfonts}
\usepackage{bbding} 
\usepackage{color}
\usepackage{amsmath}
\usepackage{CJKutf8}

\definecolor{orange}{RGB}{255,127,0}

\begin{document}
\begin{CJK}{UTF8}{gbsn}

%
\title{MSDF: A General Open-Domain Multi-Skill Dialog Framework}
%
%
\author{
Yu Zhao\inst{*}
\and
Xinshuo Hu\inst{*}
\and
Yunxin Li
\and
Baotian Hu\inst{(}\Envelope\inst{)}
\and
Dongfang Li 
\and \\
Sichao Chen
\and
Xiaolong Wang
}

\authorrunning{Y. Zhao et al.}
%
\institute{
Harbin Institute of Technology, Shenzhen \\
\email{zhaoyuhitsz@163.com
yanshekwoo@foxmail.com \\
liyunxin987@163.com 
hubaotian@hit.edu.cn \\
crazyofapple@gmail.com 
csc010228@foxmail.com \\
xlwangsz@hit.edu.cn
}
}

\maketitle              

\newcommand \footnoteONLYtext[1]
{
	\let \mybackup \thefootnote
	\let \thefootnote \relax
	\footnotetext{#1}
	\let \thefootnote \mybackup
	\let \mybackup \imareallyundefinedcommand
}
\footnoteONLYtext{*~Y. Zhao and X. Hu contribute equally to this work.}
\footnoteONLYtext{\Envelope~Corresponding author}

\input{sources/0-abstract}

\input{sources/1-introduction}

\input{sources/2-relatedwork}

\input{sources/3-method}

\input{sources/4-experiment}

\input{sources/5-conclusion}

\bibliographystyle{splncs04}

\input{sources/6-reference.bbl}
\appendix
\newpage
\input{sources/7-appendix}

\end{CJK}
\end{document}

%% file: sources/0-abstract.tex
\begin{abstract}
Dialog systems have achieved significant progress and have been widely used in various scenarios.
The previous researches mainly focused on designing dialog generation models in a single scenario, while comprehensive abilities are required to handle tasks under various scenarios in the real world. In this paper, we propose a general \textbf{M}ulti-\textbf{S}kill \textbf{D}ialog \textbf{F}ramework, namely \textbf{MSDF}, which can be applied in different dialog tasks (e.g. knowledge grounded dialog and persona based dialog).
Specifically, we propose a transferable \emph{response generator} pre-trained on diverse large-scale dialog corpora as the backbone of MSDF, consisting of BERT-based encoders and a GPT-based decoder. 
To select the response consistent with dialog history, we propose a \emph{consistency selector} trained through negative sampling.
Moreover, the flexible copy mechanism of external knowledge is also employed to enhance the utilization of multiform knowledge in various scenarios.
We conduct experiments on knowledge grounded dialog,  recommendation dialog, and persona based dialog tasks.
The experimental results indicate that our MSDF outperforms the baseline models with a large margin.
In the Multi-skill Dialog of 2021 Language and Intelligence Challenge, our general MSDF won the 3rd prize, which proves our MSDF is effective and competitive.

\end{abstract}

%% file: sources/1-introduction.tex
\section{Introduction}
\begin{figure}
\includegraphics[width=\textwidth]{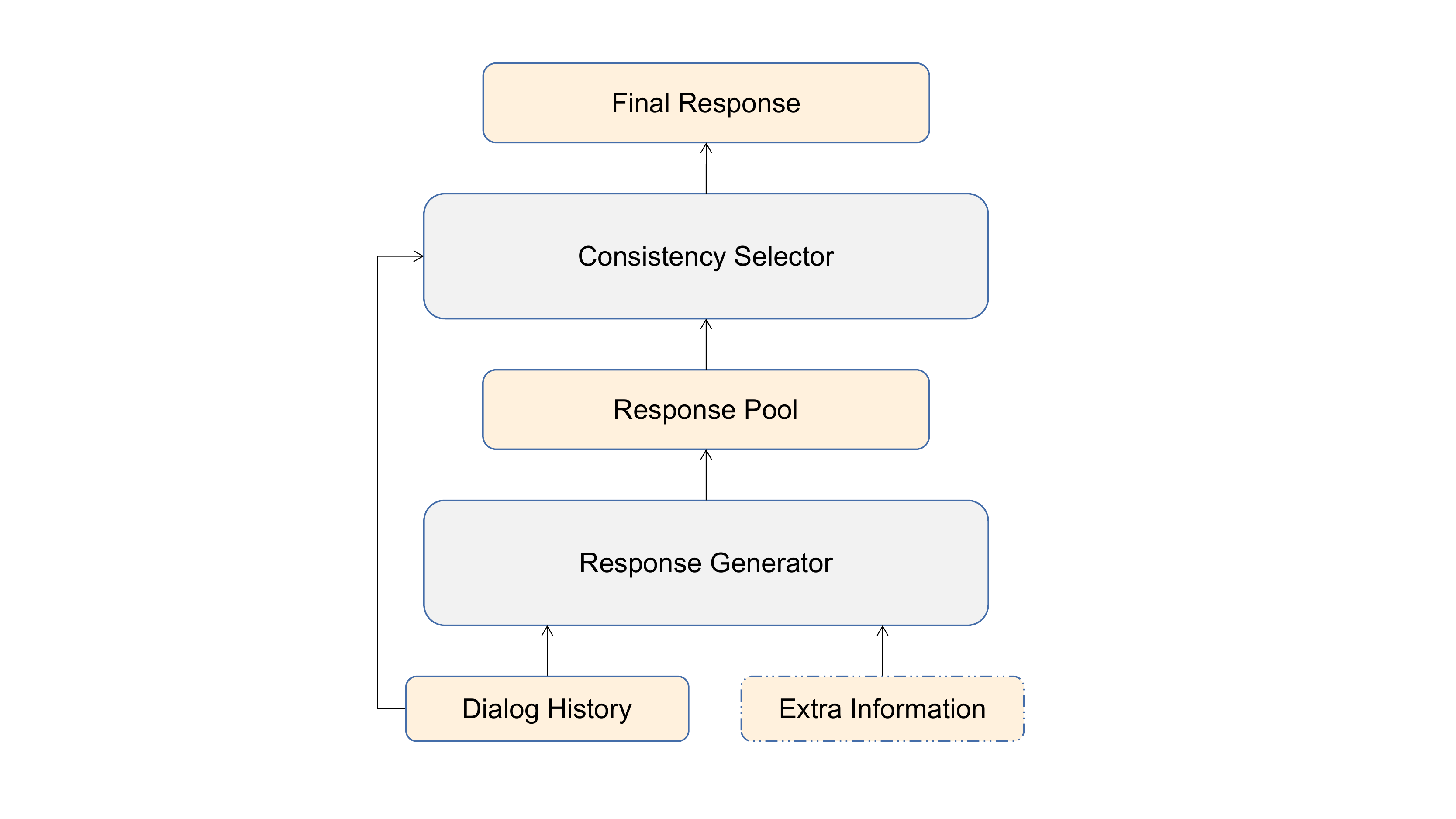}
\caption{Overall Multi-Skill Dialog Framework (MSDF). 
The response generator first generates diverse responses through a sampling-based decoding algorithm conditioned on the dialog history and optional extra information(e.g. knowledge, user profile and/or machine persona). Then, the consistency selector selects the most contextually consistent response as the final response.} \label{OverallFramework}
\end{figure}

Propelled by the acquisition of large-scale dialog corpora and the advance of pre-training technology, dialog generation models have made great progress, and dialog systems have been applied in various scenarios, such as chit-chat, knowledge grounded dialog, and conversational recommendation.
However, most of the previous works only focused on modeling the dialog system within a single scenario, which is difficult to be applied in other scenarios directly.
It does not satisfy the requirements of practical application in the real world, where the dialog model needs to generate responses in various scenarios. To handle different tasks in various scenarios, multiple dialog skills are requested, such as knowledge utilization, commodities recommendation, and persona understanding skills.
It is of great necessity to model a general multi-skill dialog framework that can be flexibly applied in various scenarios.

How to model the general multi-skill dialog systems that can effectively use information from diverse sources still remains challenging.
On the one hand, the model needs to use various information (e.g. structured and unstructured knowledge, persona, conversation topics), and the previous works usually design complex models \cite{DuRecDial}\cite{kim2019sequential} to utilize specific information in a single scenario, which results in lacking universality.
On the other hand, the previous works used complicated data processes  and training processes  \cite{liu2020you} to optimize models on specific dialog corpus, thus, the model is difficultly transferred to other scenarios.

In this work, we propose the general multi-skill dialog framework \textbf{MSDF} to address the above problems, which consists of a pre-trained dialog \emph{response generator} and a \emph{consistency selector}. As depicted in Fig.\ref{OverallFramework}, MSDF generates responses in two stages: 1) generating diverse responses as the candidate \emph{response pool}, via \emph{dialog history} (and \emph{extra information}); 2) selecting the \emph{final response} by consistency selector.
Specifically, we first pre-train a universal and transferable encoder-decoder based model on various diverse dialog corpora, including chit-chat, knowledge dialog, and recommendation dialog, to obtain a general model with multiple coarse-grained skills enhanced.
Then, we apply multiple identical encoders to encode different source information and equip the decoder with the multi-source information fusion module, which can be flexibly transferred to different application scenarios. 
Moreover, a dialog history can be mapped into multiple acceptable responses, which is also known as the one-to-many mapping~\cite{ni2021recent}, especially in the open domain. 
Thus, we introduce a BERT-based consistency selector to choose the most contextually consistent response with dialog history from the response pool, which is trained via negative sampling to distinguish consistent and inconsistent responses with dialog history.

Experiments are conducted to evaluate the performance of our MSDF in knowledge grounded dialog, recommendation dialog, and persona dialog tasks.
The experimental results indicate that our MSDF outperforms the baseline models with a large margin in terms of F1 and BLEU scores.
In the Multi-skill Dialog of 2021 Language and Intelligence Challenge, our MSDF won the third prize, with 6th rank in the human evaluation of the finals.
Both automatic and human evaluation results indicate that our MSDF is effective and competitive.

Our contributions are as bellows.
\begin{itemize}
    \item
    We propose a general multi-skill dialog framework that solves various tasks in different scenarios, namely MSDF, consisting of a dialog \emph{response generator} with multi-source information encoders and a \emph{consistency selector}.
    \item 
    We pre-train an encoder-decoder based dialog generation model on various types of large-scale open-domain dialog corpora, which can be effectively transferred into our MSDF to solve different tasks.
    \item 
    The experimental results indicate that our MSDF outperforms the baseline models with a large margin in knowledge dialog, recommendation dialog, and persona dialog tasks, demonstrating the effectiveness of MSDF.
\end{itemize}

%% file: sources/2-relatedwork.tex
\section{Related Work}
The Multi-Skill Dialog of 2021 Language and Intelligence Challenge focuses on three kinds of dialog generation tasks, including the knowledge grounded dialog, recommendation dialog, and persona dialog.
The previous works always modeled dialog systems to solve a single task in a scenario.
For knowledge grounded dialogs, the model generates responses conditioned on dialog history and the given external knowledge.
\cite{lian2019learning} employed posterior probability distribution of knowledge to guide the learning of prior distribution during training.
\cite{kim2019sequential} proposed the sequential decision making method to select knowledge used in response. And \cite{moon2019opendialkg} modeled the knowledge selection as walking over knowledge graphs.
For the persona dialogs, the model is required to understand and utilize the given persona to generate personalized responses.
\cite{liu2020you} improved the performance of persona understanding through persona negative sampling. \cite{xu2020neural} trained a VAE(Variational Auto-encoder) model to produce persona-related topic words jointly generating responses.
For recommendation dialogs, the model needs to make recommendations through conversations, which is usually separated to two tasks: recommendation and dialog generation.
\cite{chen2019towards} \cite{zhou2020improving}
incorporated the common sense knowledge graph to improve the user profile understanding and recommendation dialog generation. \cite{zhou2020towards} incorporated topic planning to enhance the recommendation dialog.
Following the work of \cite{multiGPT}, we also propose a general multi-skill dialog framework that can handle various tasks.

%% file: sources/3-method.tex
\section{Method}

\subsection{Multi-skill Dialog}
The multi-skill dialog task aims to construct a general dialog generation model with multiple skills to solve various tasks in various scenarios.
In our work, we focus on the dialog modeling task with multiple skills, including three sub-tasks: 1) knowledge dialog, 2) recommendation dialog, and 3) persona dialog.
The descriptions of the three sub-tasks are as follows\footnote{Details in: https://aistudio.baidu.com/aistudio/competition/detail/67}.

\begin{figure}
\includegraphics[width=\textwidth]{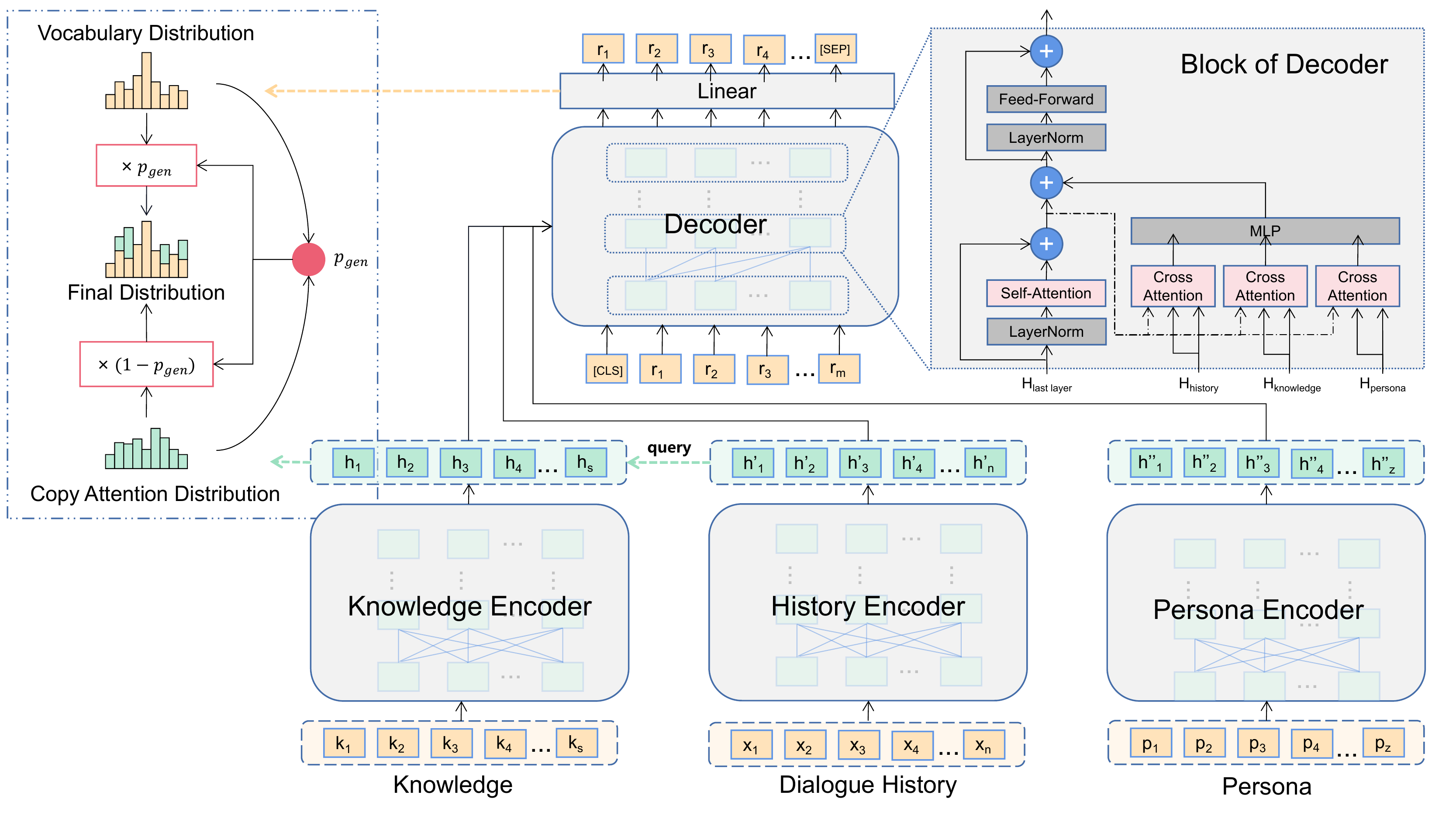}
\caption{Responses generation process in MSDF.
The top-right corner is the detail of decoder block, and the top-left corner shows the process of copy mechanism.
} \label{mBERT2GPT}
\end{figure}

\subsubsection{Knowledge Dialog}
The inputs of the knowledge dialog include dialog goals $\mathbf{G}=[\mathbf{g}_1, \mathbf{g}_2, … , \mathbf{g}_{r}] $, knowledge information $\mathbf{K}=[\mathbf{k}_1, \mathbf{k}_2, … , \mathbf{k}_{s}] $, and dialog history $ \mathbf{H}=[\mathbf{x}_1, \mathbf{x}_2, … , \mathbf{x}_{n}] $, where $\mathbf{g}_i$ is a description of dialog target or a conversation topic, $\mathbf{k}_i$ is the form of a triple or an unstructured sentence, and $\mathbf{x}_i$ consists of a sequence of words. 
The knowledge grounded dialog generation task requires the model to use appropriate knowledge to generate the response $ \mathbf{R}=[r_1, r_2, … , r_{m}] $, where $r_i$ denotes the $ i $-th word in response, $m$ denotes the length of response.

\subsubsection{Recommendation Dialog}
The inputs of the recommendation dialog include user profiles $ \mathbf{U}=[\mathbf{u}_1, \mathbf{u}_2, … , \mathbf{u}_{t}] $, conversation scenario $ \mathbf{S} $, dialog targets $\mathbf{G}$, knowledge information $\mathbf{K}$, and dialog history $ \mathbf{H}$, where $\mathbf{u}_i$ is an aspect of the user profile in a key-value form.
The recommendation dialog generation task requires the model to make recommendations based on the user profile and the scenario through conversation.

\subsubsection{Persona Dialog}
The inputs of the persona dialog include persona information of machine $ \mathbf{P}=[\mathbf{p}_1, \mathbf{p}_2, … , \mathbf{p}_{z}] $, and dialog history $ \mathbf{H}$, where $ \mathbf{p}_i$ is a sentence describing the persona.
The persona dialog generation task requires the model to generate natural, fluent and informative machine response in line with dialogue history and persona information.

\subsection{Overall Framework}

We propose a general multi-skill dialog framework MSDF, mainly consisting of a response generator and a history-response consistency selector, as shown in Fig.\ref{OverallFramework}. MSDF generates the response in two steps: response generator is applied to generate diverse responses as the response pool, and then consistency selector chooses the most consistent response from the response pool as the final output.

\subsubsection{Response Generator}
The previously proposed generation models, such as GPT2~\cite{GPT2}, PLATO2~\cite{Plato-2}, and BART~\cite{BART}, have shown their great generation performance obtaining from large-scale corpora. 
To obtain a universal pre-trained dialog model that can be flexibly transferred to the three sub-tasks, we pre-train an encoder-decoder-based model, which consists of a BERT-based encoder and a GPT-based decoder, and the optimization objective is to generate a target response conditioned on dialog history.
The six large-scale dialog datasets, including LCCC~\cite{LCCC}, Weibo\cite{WeiboDataset}, Douban\cite{DoubanDataset}, DuConv~\cite{DuConv}, KdConv~\cite{kdconv} and DuRecDial~\cite{DuRecDial}, are used to pre-train the model.
Specifically, we process all the data to a general dialog form: for the multi-turn conversations, we split them into history-response pairs; for the conversations with extra knowledge or user profiles, we ignore them during pre-training.
Moreover, to accelerate the pre-training process, we use the parameters of DialoGPT~\cite{DIALOGPT} to initialize both the encoder and decoder.
We find that the model pre-trained with all data tends to generate short responses, thus, we additionally select the half of data with longer responses to pre-train it in the latter part of pre-training phase.
In each sub-task, we duplicate the encoder of pre-trained model to encode specific information, and equip the decoder with an information fusion module. 

Denote $\mathbf{H}_{l-1}^D \in \mathbb{R}^{L_d\times d}$ as the output of ($l-1$)-th decoder block, where $L_D$ is the length of decoding sequence and $d$ is the hidden size of decoder blocks, the calculating process in each decoder block is described as follows:
\begin{equation}
\mathbf{H}_l^{SA} = SelfAttention\left(LayerNorm(\mathbf{H}_{l-1}^D)\right) + \mathbf{H}_{l-1}^D ,
\label{eq:selfattention}
\end{equation}
\begin{equation}
    \mathbf{H}_l^{FA} = Fusion\left(
        \mathbf{H}_l^{SA}, \mathbf{H}_{history},  \mathbf{H}_{knowledge}, \mathbf{H}_{persona}
    \right) ,
\label{eq:fa}
\end{equation}
\begin{equation}
\mathbf{H}_l^D=FFN
\left(
    LayerNorm (\mathbf{H}_l^{FA})
\right) + \mathbf{H}_l^{FA} ,
\label{eq:out}
\end{equation}
where $\mathbf{H}_{history} \in \mathbb{R}^{n\times d} $, $\mathbf{H}_{knowledge} \in \mathbb{R}^{s\times d} $ and $\mathbf{H}_{persona} \in \mathbb{R}^{z\times d} $ denotes the encoded information, $\mathbf{H}_l^{SA}  \in \mathbb{R}^{L_d\times d}$ denotes the result of self-attention, and $\mathbf{H}_l^{FA}  \in \mathbb{R}^{L_d\times d}$ denotes the fused multi-source information.
The $Fusion$ function is calculated in the following way:
\begin{equation} 
\mathbf{A}_l(\mathbf{H}^E) = softmax 
\left(
    \frac{ \left(LayerNorm (\mathbf{H}_l^{SA}) \mathbf{W}_l^Q \right) (\mathbf{H}^E \mathbf{W}_l^K)^T}
    {\sqrt{d}}
\right)
(\mathbf{H}^E \mathbf{W}_l^V) ,
\end{equation}

\begin{equation} \mathbf{H}_l^{FA} = [\mathbf{A}_l (\mathbf{H}_{history}); \mathbf{A}_l (\mathbf{H}_{knowledge}); \mathbf{A}_l (\mathbf{H}_{persona})] \mathbf{W}_l^P + \mathbf{H}_l^{SA} , \end{equation} 
where [;] indicates concatenation on the last dimension, $ \mathbf{W}_l^Q$, $ \mathbf{W}_l^K$ and $ \mathbf{W}_l^V$ are learnable projection matrices and $ \mathbf{W}_l^P\in \mathbb{R} ^{3d \times d} $ is a learnable attention fusing matrix (we get $ \mathbf{W}_l^P\in \mathbb{R} ^{2d \times d} $ when there are only two cross-attention to be fused). 

Finally, the hidden state $ \mathbf{H}_l^{D} \in \mathbb{R} ^{L_D \times d}$ of the $l$-th decoder layer is output by the $FFN$ layer and residual layer:
\begin{equation} \mathbf{H}_l^D=FFN(LayerNorm (\mathbf{H}_l^{FA})) + \mathbf{H}_l^{FA}. \end{equation}

With the great performance of hybrid pointer-generator network \cite{pointerNetwork} in summarization, we utilize the attention-based copy mechanism to generate knowledge-enhanced response. 
Decoder hidden states are put into linear language model to get original vocabulary distribution $ \mathbf{P}_{vocab} \in \mathbb{R}^{L_D \times L_{V}} $,
\begin{equation} \mathbf{P}_{vocab}=softmax (\mathbf{H}^D \mathbf{W}^{LM}), \end{equation} 
where $  W^{LM} \in \mathbb{R}^{d \times L_{V}} $ is the learnable language model head and $ L_{V} $ denotes the vocabulary length. 
And then attention-based copy mechanism is utilized to generate extra knowledge enhanced response. 
We obtain knowledge copy attention $ \mathbf{A}_{copy} \in \mathbb{R}^{L_D \times L_K} $, via cross-attention of decoded hidden states and  encoded knowledge (or persona) hidden state:
\begin{equation} \mathbf{A}_{copy}=softmax ((\mathbf{H}^D \mathbf{W}^Q) (\mathbf{H}_{knowledge} \mathbf{W}^K)^T), \end{equation} 
where $ \mathbf{W}^Q$ and $ \mathbf{W}^K$ are learnable projection matrices, and $ L_K $ denotes the length of the knowledge sequence. 
Then generation probability $ \mathbf{p}_{gen} \in [0,1] $ can be calculated:
\begin{equation} \mathbf{p}_{gen}=sigmoid ([\mathbf{A}_{copy} \mathbf{H}_{knowledge}; \mathbf{H^D}] \mathbf{W}^{mlp}), \end{equation} 
where $ W^{mlp} \in R^{2d \times 1} $ is learnable matrices. 
We obtain the following probability distribution over the merged vocabulary  to predict word w:
\begin{equation} \mathbf{P}(w) = \log (\mathbf{p}_{gen} \mathbf{P}_{vocab}(w) + (1- \mathbf{p}_{gen}) \mathbb{I}(w_{i}=w) A_{copy}(w) ), \end{equation}
where $ \mathbb{I}(\cdot) $ is an indicator function, and $ A^{copy}(w) $ is the element at $w$-th column in the copy attention matrix.

\subsubsection{Consistency Selector}\label{section:cs} Inspired by the BERT next sentence prediction~\cite{BERT}, we propose a history-response consistency selector, since the response should be the next sentence by dialog history. The consistency selector is a binary classifier to distinguish the consistent and inconsistent response with dialog history. We construct the consistency selector by a pre-trained model RoBBERTa~\cite{Roberta}, plus a linear head layer. We first concatenate the dialog history and response as the pre-trained model input to get context representation, and then put the first token (usually known as [CLS] token) to the linear head to get a binary classification score. The consistency selector is trained on positive and negative sampling examples from training data, where inconsistent responses are randomly sampled. During inference, we get the positive score from the consistency selector output as the consistency score.

\subsection{Data Processing}

We separate different input resources into four categories: dialog history, knowledge information, persona information, and current reply (or previously generated response during inference). 
The data preprocessing and reprocessing are demonstrated as follows.

The knowledge dialog generation model consists of a dialog history encoder, a knowledge encoder and a decoder. During data preprocessing of DuConv, we reformat the knowledge graph to a pseudo unstructured knowledge sentences, by concatenating the subject, predicate and object, and join all knowledge sentence with special token ``[SEP]". Since the dialog topics are limited in movies and film stars and there are at most two topics during a dialog, we introduce special tokens to format the subject in the knowledge graph, including ``[movie1]", ``[movie2]", ``[star1]", and ``[star2]", which will be restored in data reprocessing after response generated. And we also add speaker tokens ``[speaker1]" and ``[speaker2]" to distinguish speakers in the dialog history. We also reprocess personal pronouns and figures due to the knowledge, such as outcome date of movies, and birthday of stars.

The recommendation dialog generation model consists of a dialog history encoder, a knowledge encoder, a persona encoder and a decoder. To facilitate optimization, we make the knowledge encoder and persona encoder share parameters. The knowledge graph in DuRecDial is also reformatted in the same way as DuConv, with all the candidate recommendation goal planning concatenated at the end. Since the recommendation goal is labeled with each response, we introduce a new special token ``[goal]" as a separator, and join the golden goal with response， e.g. ``[goal]问User性别[goal]我该称呼您是先生还是女士"(``[goal] Asking about the user's gender [goal] Should I call you Mr. or MS"). This golden goal prefix in the response performs as the semantic guidance, which is typically like conditional generation. The dialog situation is viewed as extra knowledge concatenated before knowledge information. We also replace the user name with a new special token ``[uname]" and add speaker tokens. Personal pronouns and figures will be corrected and goal information will be removed during data reprocess. 

The persona dialog generation model consists of a dialog history encoder, a persona encoder and a decoder. We trained a simple Word2Vec by open-source gensim implementation \footnote{https://radimrehurek.com/gensim/intro.html}, on CPC sentences to estimate the similarity between the  response and persona by cosine similarity of average word vector, and randomly drop training examples with persona information similarity less than 0.7. It is observed that the generator prefers to generate longer sentences and there are too many responses whose topic is about the user job, so we drop some examples with too long responses and ``工作"(``work") in dialog utterances. Considering that conversations in CPC are more like chit-chat than task-oriented dialog, we abandon the copy mechanism in this generator.

%% file: sources/4-experiment.tex
\section{Experiment}
\subsection{Experimental Settings}

\subsubsection{Dataset} For pre-training, we use the Weibo dataset~\cite{WeiboDataset}, Douban multi-round conversation~\cite{DoubanDataset}, LCCC dataset~\cite{LCCC}, Emotional Conversational Dataset~\cite{EmotionalDataset}, retrieval-assisted conversational dataset~\cite{RetrievalDataset}, and Kdconv dataset~\cite{kdconv}. For fine-tuning, we use DuConv~\cite{DuConv} for knowledge grounded dialog, DuRecDial~\cite{DuRecDial} for the conversational recommendation, and Chinese persona chat (CPC) for persona dialog. 
Statistics of all dataset are summarized in Table.~\ref{datasets}. 

\input{tables/datasets}

\subsubsection{Implementation Details} For the response generator, the encoders and decoder settings follow the DialoGPT~\cite{DIALOGPT}, where the hidden size is 768 for 12 layers, the maximum input length is 512, and there are 12 heads in multi-head attention. For all dropout layers, the dropout rate is set to 0.1. 
We adopt cross-entropy loss
as our loss function, and the parameters are saved in term of the minimum cross-entropy loss on the development datasets. The consistency selector is implemented alike a NLI model as described in section~\ref{section:cs},  which is also optimized through cross-entropy loss function.

\subsection{Evaluation}

The competitive baseline models are compared with the proposed MSDF on different quantitative metrics. We use BLEU-1 and BLEU-2 to evaluate the n-gram lexical similarity, F1 to evaluate the Chinese character level similarity, and DISTINCT1 and DISTINCT-2 to evaluate the diversity of generated responses. The total SCORE for the automatic evaluation is calculated by averaging all the F1/BLEU1/BLEU2 scores for the subtasks. To exhibit the improvement of our MSDF, we also implement and test four baseline models: Seq2Seq~\cite{seq2seq}, HRED~\cite{hred}, DialoGPT~\cite{DIALOGPT} and BERT-GPT~\cite{BertGPT}. All these baseline models only take dialog history as inputs, without aquiring extra information. Since test datasets are not released, we evaluate all the models on LUGE platform~\footnote{LUGE: https://aistudio.baidu.com/aistudio/competition/detail/48}. The performance of baseline models and our MSDF is presented in Table.~\ref{AutoEval}, including ablation experiments.

With respect to the human evaluation, we refer readers to the competition leaderboard~\footnote{Leaderboard: https://aistudio.baidu.com/aistudio/competition/detail/67} for the single-turn or multi-turn dialog evaluation results.

\input{tables/AutoEval}

\subsection{Discussion} 
Our MSDF outperforms the baseline models with a large margin, even though without extra information, which strongly presents the effectiveness. The consistency selector preferred to select the most common response. It reduces the variance of generating performance and significantly improves automatic evaluation results, without considering the limited diversity of responses in human evaluation. 
Besides, the attention-based copy mechanism is of great importance for generating knowledge-enhanced response.
According to our observation, DialoGPT and BERT-GPT still talk rubbish after fine-tuning, despite resulting in higher DISTINCT scores.
In the competition, our MSDF got 9-th rank in automatic evaluation and 6-th rank in human evaluation.
It increased by 3 ranks in human evaluation compared to the automatic evaluation. We attribute this to the multi-skill pre-training, from which our model could generate more human-like responses, in spite of the mismatch with the golden references in automatic evaluation. 

%% file: tables/datasets.tex
\begin{table}
\centering
\caption{Statistics of all datasets. $^{\dag}$ denotes the datasets for fine-tuning, and the others are for pre-training.}\label{datasets}
\begin{tabular}{p{6.2cm}|p{1.8cm}|p{1.8cm}|p{1.8cm}}
\hline
 & \bfseries Train & \bfseries Dev & \bfseries Test  \\
\hline
DuConv$^{\dag}$~\cite{DuConv} & 19858 & 2000 & 5000 \\
DuRecDial$^{\dag}$~\cite{DuRecDial} & 6618 & 946 & 4645 \\
Chinese Persona Chat$^{\dag}$ & 23000 & 1500 & 3000 \\
Weibo dataset~\cite{WeiboDataset}  & 3103764 & 443394 & 886790 \\
Douban multi-round conversation~\cite{DoubanDataset} & 5000000 & 25001 & 1186 \\
LCCC dataset~\cite{LCCC} & 11987759 & 20000 & 10000 \\
Emotional Conversational Dataset~\cite{EmotionalDataset} & 899207 & 110000 & 110000 \\
retrieval-assisted conversational dataset~\cite{RetrievalDataset} & 5498480 & 107332 & 156706 \\
Kdconv dataset~\cite{kdconv} & 3000 & 300 & 2751 \\
\hline
\end{tabular}
\end{table}

%% file: tables/AutoEval.tex
\begin{table}
\centering
\caption{Automatic evaluation of test set B on baseline models and our MSDF (ranked by total score). All the results of F1, BLEU, and DISTINCT are average scores from three sub-tasks.}\label{AutoEval}
\begin{tabular}{p{3.2cm}|p{1.6cm}|p{1.2cm}|p{2.6cm}|p{2.6cm}}
\hline
\bfseries  & \bfseries SCORE & \bfseries F1 & \bfseries BLEU1/2 & \bfseries DISTINCT1/2 \\
\hline
Seq2Seq~\cite{seq2seq}     & 0.522 & 22.33 & 0.202/0.096 & 0.038/0.100 \\
HRED~\cite{hred}        & 0.565 & 23.66 & 0.220/0.108 & 0.038/0.105 \\
DialoGPT~\cite{DIALOGPT}   & 0.373 & 17.05 & 0.141/0.061 & {\bfseries 0.079/0.313} \\
BERT-GPT~\cite{BertGPT} & 0.573 & 24.51 & 0.215/0.113 & 0.061/0.214 \\
\hline
MSDF & {\bfseries 0.934}	& {\bfseries 38.62}	& {\bfseries 0.333/0.215} & 0.057/0.183 \\
~ -consistency selector & 0.872 & 36.00 & 0.314/0.198 & 0.051/0.173 \\
~ -extra knowledge & 0.705 & 28.98 & 0.271/0.145 & 0.049/0.162 \\
\hline
\end{tabular}
\end{table}

%% file: sources/5-conclusion.tex
\section{Conclusion}

This paper describes our general multi-skill dialog framework MSDF, consisting of a response generator and a dialog history consistency selector. We first pre-train the basic encoder-decoder on diverse datasets and then fine-tune it on the specific dataset to construct the response generator with a strong specific skill. We won the third prize in Multi-task Dialog of 2021 Language and Intelligence Challenge. Experiments are also conducted on several baseline models. The vast improvement over the baseline models indicates that our framework is effective and competitive. In future work, we will experiment with our MSDF on more tasks to evaluate the comprehensive skills of our framework and further improve its performance.

\section*{Acknowledgement}
\begin{sloppypar}
We appreciate the insightful feedback from the anonymous reviewers and the beneficial support from Baidu Inc. This work is jointly supported by grants: Natural Science Foundation of China (No. 62006061 and 61872113), Strategic Emerging Industry Development Special Funds of Shenzhen (JCYJ20200109113403826 and JCYJ20200109113441941) and Stable Support Program for Higher Education Institutions of Shenzhen (No. GXWD20201230155427003-20200824155011001).
\end{sloppypar}

%% file: sources/7-appendix.tex
\section{Experimental Setting}
\label{sec:Setting}
\input{tables/ExperimentalSetting}
\newpage

\section{Samples}
\label{sec:CaseStudy}
\input{tables/duconvCase}
\newpage
\input{tables/personaCase}
\newpage
\input{tables/durecdialCase}

%% file: tables/ExperimentalSetting.tex
\begin{table}[h]
\caption{The experimental setting details for each task.}\label{Settings}
\begin{tabular}{l|l}
\hline
\multicolumn{2}{l}{{\bfseries Knowledge Dialog}} \\
\hline
Architecture    &   History encoder, Knowledge encoder, Decoder \\
Learning rate   &   $ 2.5 \times 10^{-5} $ \\
Batch size      &   $ 16 $ \\
Max epochs      &   $ 15 $ \\
Optimizer       &   AdamW($ \beta _1 = 0.9, \beta _2 = 0.999, \epsilon = 10 \times 10^{-6} $) \\
Copy mechanism  &   Knowledge copy mechanism \\
\hline
\multicolumn{2}{l}{{\bfseries Recommendation Dialog}} \\
\hline
Architecture    &   History encoder, Knowledge encoder, Persona encoder, Decoder \\
Learning rate   &   $ 2.5 \times 10^{-5} $ \\
Batch size      &   $ 2 $ \\
Max epochs      &   $ 10 $ \\
Optimizer       &   AdamW($ \beta _1 = 0.9, \beta _2 = 0.999, \epsilon = 10 \times 10^{-6} $) \\
Copy mechanism  &   Knowledge copy mechanism \\
\hline
\multicolumn{2}{l}{{\bfseries Persona Dialog}} \\
\hline
Architecture    &   History encoder, Persona encoder, Decoder \\
Learning rate   &   $ 2.5 \times 10^{-5} $ \\
Batch size      &   $ 16 $ \\
Max epochs      &   $ 10 $ \\
Optimizer       &   AdamW($ \beta _1 = 0.9, \beta _2 = 0.999, \epsilon = 10 \times 10^{-6} $) \\
Copy mechanism  &   No copy mechanism \\
\hline
\end{tabular}
\end{table}

%% file: tables/duconvCase.tex
\begin{table}[h]
\scriptsize
\centering
\caption{Example of multi-turn knowledge dialog generation results. Due to the topic formatting, the generator performs well in capturing knowledge graph subject and object information.}\label{duconvCase}
\renewcommand\arraystretch{1.3}
\begin{tabular}{|l|p{11cm}|}
\hline
Knowledge graph &
\multicolumn{1}{m{11cm}|}{
[{\bfseries 碟中谍},国家,美国]\newline
(Mission: Impossible, country, USA)\newline
[{\bfseries 碟中谍},口碑,口碑一般]\newline
(Mission: Impossible, public praise, common)\newline
[\textcolor{cyan}{{\bfseries 碟中谍},类型,动作]} \newline
(Mission: Impossible, type, action)\newline
[{\bfseries 碟中谍},领域,电影]\newline
(Mission: Impossible, field, movie)\newline
[\textcolor{red}{{\bfseries 碟中谍},主演,强·沃特]} \newline
(Mission: Impossible, starring, Jon Voight)\newline
[{\bfseries 碟中谍},幕后花絮,电视剧版《谍中谍》的铁杆粉丝非常不满本片对吉姆的 描写]\newline
(Mission: Impossible, behind the scenes, the hardcore fans of the TV series “Mission: Impossible” are very dissatisfied with the film's description of Jim. )\newline
[{\bfseries 碟中谍},时光网评分,7.9]\newline
(Mission: Impossible, Mtime score, 7.9)\newline
[{\bfseries 强·沃特},评论,又一老色鬼]\newline
(Jon Voight, comment, another old pervert)\newline
[\textcolor{red}{{\bfseries 强·沃特},星座,摩羯座]} \newline
(Jon Voight, constellation, capricorn)\newline
[{\bfseries 强·沃特},身高,191cm]\newline
(Jon Voight, Height, 191cm)\newline
[{\bfseries 强·沃特},性别,男]\newline
(Jon Voight, gender, male )\newline
[{\bfseries 强·沃特},职业,演员]\newline
(Jon Voight, occupation, actor)\newline
[{\bfseries 强·沃特},领域,明星] \newline
(Jon Voight, field, star)\newline }\\ 
\hline
Conversation &
\multicolumn{1}{m{11cm}|}{
   {\bfseries Bot}: 你喜欢看什么类型的电影？ \newline
   (What kind of movies do you like to watch?)\newline
   {\bfseries User}: \textcolor{cyan}{动作类} \newline
   (Action)\newline
   {\bfseries Bot}: 给你推荐一部电影，名字叫\textcolor{cyan}{碟中谍} ， 有时间可以看一下。 \newline  
   (I'd like to recommend a movie called “Mission: Impossible”. You can watch it when you have time.)\newline
   {\bfseries User}: 好看吗 \newline 
   (Does it look good)\newline
   {\bfseries Bot}: \textcolor{red}{这部电影的主演}跟你一样都是\textcolor{red}{摩羯座}的，你知道是谁吗？ \newline 
   (The starring actors in this movie are Capricorns just like you. Do you know who it is?)\newline
   {\bfseries User}: 是谁呀 \newline  
   (Who is it)\newline
   {\bfseries Bot}: \textcolor{red}{强·沃特，摩羯座}的，你可以关注一下。  \newline
   (Jon Voight, Capricorn, you can pay attention to him.) 
   }
 \\ \hline
\end{tabular}
\end{table}

%% file: tables/personaCase.tex
\begin{table}[h]
\scriptsize
\centering
\caption{Example of multi-turn persona dialog generation results. The generation model can generate some persona-based response, while most of the time replied sentences are uncorrelated with persona, and the consistency selector prefers common response. We lay the blame on the open-domain dataset, although we filter training examples with more persona correlation.}\label{personaCase}
\renewcommand\arraystretch{1.3}
\begin{tabular}{|l|p{11cm}|}
\hline
Bot persona &
\multicolumn{1}{m{11cm}|}{
{\bfseries persona0}: \textcolor{cyan}{我最喜欢的歌手是毛不易}\newline (My favorite singer is Buyi Mao)\newline
{\bfseries persona1}: 我真的很喜欢看《极限挑战》\newline
(I really like watching “Extreme Challenge”)\newline
{\bfseries persona2}: 我喜欢吃脆饼干 \newline  
(I like to eat crispy biscuits) \newline
{\bfseries persona3}: \textcolor{red}{我最喜欢的乐队是五月天}\newline
(My favorite band is Mayday)
} \\ \hline
Conversation &
\multicolumn{1}{m{11cm}|}{
  {\bfseries User}: 哈哈，是的还是第一部最好看了。\newline 
  (Ha ha, yes, the first one is the best to watch.)\newline
  {\bfseries Bot}: \textcolor{red}{是啊，我最喜欢的乐队是五月天}，你知道这个乐队 吗？\newline
  (Yes, my favorite band is Mayday. Do you know this band?)\newline
  {\bfseries User}: 我知道啊，我挺喜欢这个乐队的\newline
  (I know. I really like this band.)\newline
  {\bfseries Bot}: 是吗？我也挺喜欢他们的，不过我更喜欢的是他们的歌 。\newline 
  (Is it?  I also like them very much, but I prefer their songs.)\newline
  {\bfseries User}: 他们的歌很好听\newline
  (Their songs are very nice)\newline
  {\bfseries Bot}: 是的，他们唱的歌都很好听，你有喜欢的歌手吗 ？\newline
  (Yes, they sing songs very well. Do you have any singers you like?)\newline
  {\bfseries User}: 我喜欢毛不易\newline
  (I like Buyi Mao)\newline
  {\bfseries Bot}: \textcolor{cyan}{是啊，毛不易的歌我都很喜欢。}\newline 
  (Yes, I like all Mao Buyi's songs very much.)\newline
  {\bfseries User}: 他唱歌很有感觉\newline 
  (He sings very well)\newline
  {\bfseries Bot}: 是的，我也是这么认为的，对了，你是做什么工作的啊？\newline
  (Yes, I think so. By the way, what do you do?)\newline
  {\bfseries User}: 我是一名老师\newline 
  (I am a teacher)\newline
  {\bfseries Bot}: 老师也挺好的，可以教学生知识。\newline 
  (The teacher is also very good and can teach students knowledge. )\newline
  {\bfseries User}: 是的啊\newline 
  (Yes)\newline
  {\bfseries Bot}: 是啊，你平时喜欢做些什么啊？\newline 
  (Yes, what do you like to do?)\newline
  {\bfseries User}: 看看综艺，听听音乐啥的\newline 
  (Watch variety shows, listen to music or something)\newline
  {\bfseries Bot}: 综艺节目，《向往的生活》，你看过吗？\newline
  (Variety show, “The Longing for Life”, have you ever watched it?)
  }
 \\ \hline
\end{tabular}
\end{table}

%% file: tables/durecdialCase.tex
\begin{table}[ht]
\scriptsize
\caption{Example of multi-turn recommendation dialog generation results. The generated responses are highly correlative with extra knowledge information. Note that the bot also follows the goal planning in this example, which we attribute to the goal joining ahead responses. 
}\label{durecdialCase}
\renewcommand\arraystretch{1.3}
\begin{tabular}{|l|p{11cm}|}
\hline
Situation &
\multicolumn{1}{m{11cm}|}{
\textcolor{blue}{聊天日期:2018-5-6}，聊天时间:中午12:00，在学校 \newline
(Chat date: 2018-5-6, Chat time: 12:00 noon, at school ) 
}
\\ 
\hline
Goal planning & \multicolumn{1}{m{11cm}|}{{\bfseries [1] 问日期} (User主动问 日期，Bot根据『参考知识』的『聊天日期』回答，然后User满足并好评) \newline
...... \newline 
{\bfseries [3] 电影推荐} (Bot主动，使用『娜娜的玫瑰战争』的某个评论当做推荐理由来推荐『娜娜的玫瑰战争』，User先问电影『国家地区、导演、类型、主演、口碑、评分』中的一个或多个，Bot回答，最终User接受) \newline
{\bfseries [4]再见}\newline
([1] Ask the date(User actively asks the date, Bot answers according to the "chat date" of "Reference Knowledge", and then User is satisfied and praised)  ......  [3] Movie recommendation (Bot takes the initiative , Use a comment of "Nana's War of the Roses" as a recommendation reason to recommend "Nana's War of the Roses", the user first asks one or more of the movie "country, director, genre, starring, public praise, rating" , Bot answers, and finally User accepts) --> [4] Goodbye )
} \\ 
\hline
User profile &
\multicolumn{1}{m{11cm}|}{
    {\bfseries 职业状态}:  学生 
    (Occupation status: students) \newline
    {\bfseries 年龄区间}:  小于18 
    (Age range: less than 18) \newline
    ...... \newline
    {\bfseries 喜欢的电影}: 笑功震武林 
    (Favorite movies: Princess and Seven Kung Fu Masters) \newline
    {\bfseries 拒绝}: 音乐 
    (Rejection: music) \newline
    \textcolor{orange}{{\bfseries 喜欢的明星}: 谢娜}  
    (Favorite star: Nana) \newline
    {\bfseries 接受的电影}: 大玩家 
    (Accepted movies: Super Player) \newline
    {\bfseries 同意的新闻}: 谢娜的新闻 
    (Agreed news: news about Nana) 
} \\ 
\hline
Knowledge graph &
\multicolumn{1}{m{11cm}|}{
  \textcolor{blue}{[ {\bfseries 聊天}, 日期,  2018-5-6 ]} 
  (Chat, date, 2018-5-6) \newline
  \textcolor{magenta}{[ {\bfseries 谢娜}, 生日,  1981-5-6 ]}
  (Nana, birthday, May 6, 1981) \newline
  [{\bfseries 谢娜},主演,娜娜的玫瑰战争]
  (Nana, starring, Nana's War of the Roses) \newline
  ...... \newline
  [{\bfseries 娜娜的玫瑰战争},国家地区,中国]
  (Nana's War of the Roses, country region, China) \newline
  \textcolor{cyan}{[ {\bfseries 娜娜的玫瑰战争},类型,爱情喜剧]}
  (Nana's War of the Roses, type, romantic comedy) \newline
  [{\bfseries 娜娜的玫瑰战争},主演,谢娜、李承铉、金伊彤]
  (Nana's War of the Roses, starring, Nana、Lee Seung Hyun、Yitong Jin) \newline
  \textcolor{red}{[ {\bfseries 娜娜的玫瑰战争},评论,《娜娜的玫瑰战争》不仅囊括种种恋爱难题，更通过谢娜的本色表演以及剧中的“痴心宝马男”李承铉帅炫造型，将生活中极为常见的爱情桥段以诙谐搞笑的形式释放出来。}借力片中颇耐人寻味的爱情对白，和“拿得起放得下”的豁达爱情态度，宣告新时代年轻人“爱的保卫战”开始。 
  ]
  (Nana's War of the Roses, comments, “Nana's War of the Roses” not only covers all kinds of love problems, but also releases the most common love scenes in life in the form of humor through the performance of Xie Na's true colors and the "infatuated BMW man" Lee Seung Hyun's handsome and dazzling style With the thought-provoking love dialogue and the open-minded love attitude of "can afford to put it down", the film announces the beginning of the "battle of love" for young people in the new era.) 
  } \\ 
\hline
Conversation &
\multicolumn{1}{m{11cm}|}{
   {\bfseries User}: 你好今天几号了啊？  (Hello, what's the date today?) \newline
   {\bfseries Bot}: 今天是\textcolor{blue}{2018年5月6日}。  (Today is May 6, 2018.) \newline
   {\bfseries User}: 知道啦，谢谢你 
   (I see. Thank you.) \newline
   {\bfseries Bot}: 不客气，你知道吗？今天也是\textcolor{orange}{谢娜}的生日哦，她\textcolor{magenta}{是1986年5月6日}出生的 。
   (You're welcome, you know? Today is also the birthday of Nana. She was born on May 6th, 1986.) \newline
   {\bfseries User}: 哇，真的吗，我最喜欢谢娜了
   (Wow, really? I like Nana best) \newline
   {\bfseries Bot}: \textcolor{red}{那我给你推荐一部她的电影吧，『娜娜的玫瑰战争』，不仅囊括种种恋爱难题，更通过谢娜的本色表演以及剧中的“痴心宝马男”李承铉 帅炫造型，将生活中极为常见的爱情桥段以诙谐搞笑的形式释放出来。}
   (I'd like to recommend her film, “Nana's War of the Roses”, which not only covers all kinds of love problems, but also releases the most common love scenes in life in a humorous and funny way through the natural performance of Nana and the "infatuated BMW man" Lee Seung Hyun's handsome and dazzling style in the play.) \newline
   {\bfseries User}: 这是什么类型的电影啊？ 
   (What kind of movie is this?) \newline
   {\bfseries Bot}: \textcolor{cyan}{是一部爱情喜剧片}哦 。 
   (It's a romantic comedy.) \newline
   {\bfseries User}: 我喜欢看爱情喜剧片，那我就看这个了
   (I like romantic comedies, so I will watch it.) \newline
   {\bfseries Bot}: 你喜欢就好。
   (I'm glad you like it.)
   }
 \\ \hline
\end{tabular}
\end{table}

%% file: main.bbl
\begin{thebibliography}{10}
\providecommand{\url}[1]{\texttt{#1}}
\providecommand{\urlprefix}{URL }
\providecommand{\doi}[1]{https://doi.org/#1}

\bibitem{Plato-2}
Bao, S., He, H., Wang, F., Wu, H., Wang, H., Wu, W., Guo, Z., Liu, Z., Xu, X.:
  Plato-2: Towards building an open-domain chatbot via curriculum learning.
  arXiv preprint arXiv:2006.16779  (2020)

\bibitem{RetrievalDataset}
Cai, D., Wang, Y., Bi, W., Tu, Z., Liu, X., Shi, S.: Retrieval-guided dialogue
  response generation via a matching-to-generation framework. In: Proceedings
  of the 2019 Conference on Empirical Methods in Natural Language Processing
  and the 9th International Joint Conference on Natural Language Processing
  (EMNLP-IJCNLP). pp. 1866--1875 (2019)

\bibitem{multiGPT}
Cao, Y., Bi, W., Fang, M., Tao, D.: Pretrained language models for dialogue
  generation with multiple input sources. In: Proceedings of the 2020
  Conference on Empirical Methods in Natural Language Processing: Findings. pp.
  909--917 (2020)

\bibitem{chen2019towards}
Chen, Q., Lin, J., Zhang, Y., Ding, M., Cen, Y., Yang, H., Tang, J.: Towards
  knowledge-based recommender dialog system. In: Proceedings of the 2019
  Conference on Empirical Methods in Natural Language Processing and the 9th
  International Joint Conference on Natural Language Processing (EMNLP-IJCNLP).
  pp. 1803--1813 (2019)

\bibitem{BERT}
Devlin, J., Chang, M.W., Lee, K., Toutanova, K.: Bert: Pre-training of deep
  bidirectional transformers for language understanding. In: Proceedings of the
  2019 Conference of the North American Chapter of the Association for
  Computational Linguistics: Human Language Technologies, Volume 1 (Long and
  Short Papers). pp. 4171--4186 (2019)

\bibitem{kim2019sequential}
Kim, B., Ahn, J., Kim, G.: Sequential latent knowledge selection for
  knowledge-grounded dialogue. In: International Conference on Learning
  Representations (2019)

\bibitem{BART}
Lewis, M., Liu, Y., Goyal, N., Ghazvininejad, M., Mohamed, A., Levy, O.,
  Stoyanov, V., Zettlemoyer, L.: Bart: Denoising sequence-to-sequence
  pre-training for natural language generation, translation, and comprehension.
  In: Proceedings of the 58th Annual Meeting of the Association for
  Computational Linguistics. pp. 7871--7880 (2020)

\bibitem{lian2019learning}
Lian, R., Xie, M., Wang, F., Peng, J., Wu, H.: Learning to select knowledge for
  response generation in dialog systems. In: IJCAI International Joint
  Conference on Artificial Intelligence. p.~5081 (2019)

\bibitem{liu2020you}
Liu, Q., Chen, Y., Chen, B., Jian-Guang, L., Chen, Z., Zhou, B., Zhang, D.: You
  impress me: Dialogue generation via mutual persona perception. In:
  Proceedings of the 58th Annual Meeting of the Association for Computational
  Linguistics. pp. 1417--1427 (2020)

\bibitem{Roberta}
Liu, Y., Ott, M., Goyal, N., Du, J., Joshi, M., Chen, D., Levy, O., Lewis, M.,
  Zettlemoyer, L., Stoyanov, V.: Roberta: A robustly optimized bert pretraining
  approach. arXiv preprint arXiv:1907.11692  (2019)

\bibitem{DuRecDial}
Liu, Z., Wang, H., Niu, Z.Y., Wu, H., Che, W., Liu, T.: Towards conversational
  recommendation over multi-type dialogs. In: Proceedings of the 58th Annual
  Meeting of the Association for Computational Linguistics. pp. 1036--1049
  (2020)

\bibitem{moon2019opendialkg}
Moon, S., Shah, P., Kumar, A., Subba, R.: Opendialkg: Explainable
  conversational reasoning with attention-based walks over knowledge graphs.
  In: Proceedings of the 57th Annual Meeting of the Association for
  Computational Linguistics. pp. 845--854 (2019)

\bibitem{ni2021recent}
Ni, J., Young, T., Pandelea, V., Xue, F., Adiga, V., Cambria, E.: Recent
  advances in deep learning-based dialogue systems. arXiv preprint
  arXiv:2105.04387  (2021)

\bibitem{GPT2}
Radford, A., Wu, J., Child, R., Luan, D., Amodei, D., Sutskever, I.: Language
  models are unsupervised multitask learners. OpenAI blog  \textbf{1}(8), ~9
  (2019)

\bibitem{pointerNetwork}
See, A., Liu, P.J., Manning, C.D.: Get to the point: Summarization with
  pointer-generator networks. In: Proceedings of the 55th Annual Meeting of the
  Association for Computational Linguistics (Volume 1: Long Papers). pp.
  1073--1083 (2017)

\bibitem{hred}
Serban, I.V., Sordoni, A., Bengio, Y., Courville, A., Pineau, J.: Building
  end-to-end dialogue systems using generative hierarchical neural network
  models. In: Thirtieth AAAI Conference on Artificial Intelligence (2016)

\bibitem{WeiboDataset}
Shang, L., Lu, Z., Li, H.: Neural responding machine for short-text
  conversation. In: Proceedings of the 53rd Annual Meeting of the Association
  for Computational Linguistics and the 7th International Joint Conference on
  Natural Language Processing (Volume 1: Long Papers). pp. 1577--1586 (2015)

\bibitem{seq2seq}
Sutskever, I., Vinyals, O., Le, Q.V.: Sequence to sequence learning with neural
  networks. Advances in Neural Information Processing Systems  \textbf{27},
  3104--3112 (2014)

\bibitem{LCCC}
Wang, Y., Ke, P., Zheng, Y., Huang, K., Jiang, Y., Zhu, X., Huang, M.: A
  large-scale chinese short-text conversation dataset. In: CCF International
  Conference on Natural Language Processing and Chinese Computing. pp. 91--103.
  Springer (2020)

\bibitem{DuConv}
Wu, W., Guo, Z., Zhou, X., Wu, H., Zhang, X., Lian, R., Wang, H.: Proactive
  human-machine conversation with explicit conversation goal. In: Proceedings
  of the 57th Annual Meeting of the Association for Computational Linguistics.
  pp. 3794--3804 (2019)

\bibitem{DoubanDataset}
Wu, Y., Wu, W., Xing, C., Zhou, M., Li, Z.: Sequential matching network: A new
  architecture for multi-turn response selection in retrieval-based chatbots.
  In: Proceedings of the 55th Annual Meeting of the Association for
  Computational Linguistics (Volume 1: Long Papers). pp. 496--505 (2017)

\bibitem{xu2020neural}
Xu, M., Li, P., Yang, H., Ren, P., Ren, Z., Chen, Z., Ma, J.: A neural topical
  expansion framework for unstructured persona-oriented dialogue generation.
  arXiv preprint arXiv:2002.02153  (2020)

\bibitem{BertGPT}
Zeng, G., Yang, W., Ju, Z., Yang, Y., Wang, S., Zhang, R., Zhou, M., Zeng, J.,
  Dong, X., Zhang, R., et~al.: Meddialog: A large-scale medical dialogue
  dataset. In: Proceedings of the 2020 Conference on Empirical Methods in
  Natural Language Processing (EMNLP). pp. 9241--9250 (2020)

\bibitem{DIALOGPT}
Zhang, Y., Sun, S., Galley, M., Chen, Y.C., Brockett, C., Gao, X., Gao, J.,
  Liu, J., Dolan, W.B.: Dialogpt: Large-scale generative pre-training for
  conversational response generation. In: Proceedings of the 58th Annual
  Meeting of the Association for Computational Linguistics: System
  Demonstrations. pp. 270--278 (2020)

\bibitem{EmotionalDataset}
Zhou, H., Huang, M., Zhang, T., Zhu, X., Liu, B.: Emotional chatting machine:
  Emotional conversation generation with internal and external memory. In:
  Thirty-Second AAAI Conference on Artificial Intelligence (2018)

\bibitem{kdconv}
Zhou, H., Zheng, C., Huang, K., Huang, M., Zhu, X.: Kdconv: A chinese
  multi-domain dialogue dataset towards multi-turn knowledge-driven
  conversation. In: Proceedings of the 58th Annual Meeting of the Association
  for Computational Linguistics. pp. 7098--7108 (2020)

\bibitem{zhou2020improving}
Zhou, K., Zhao, W.X., Bian, S., Zhou, Y., Wen, J.R., Yu, J.: Improving
  conversational recommender systems via knowledge graph based semantic fusion.
  In: Proceedings of the 26th ACM SIGKDD International Conference on Knowledge
  Discovery \& Data Mining. pp. 1006--1014 (2020)

\bibitem{zhou2020towards}
Zhou, K., Zhou, Y., Zhao, W.X., Wang, X., Wen, J.R.: Towards topic-guided
  conversational recommender system. In: Proceedings of the 28th International
  Conference on Computational Linguistics. pp. 4128--4139 (2020)

\end{thebibliography}
